\documentclass[sigconf]{aamas} 

\usepackage{amsmath}

\usepackage{amssymb}
\usepackage{mathtools}
\usepackage{amsthm}
\usepackage{ulem}
\usepackage{multirow}

\theoremstyle{definition}

\theoremstyle{plain} 
\newtheorem{theorem}{Theorem}

\usepackage{graphicx} 
\usepackage{subfigure}
\usepackage{subcaption}

\usepackage{algorithm}
\usepackage{amssymb}
\usepackage{pifont}

\usepackage{amsfonts}       
\usepackage{algorithmic} 
\usepackage{wrapfig}
\usepackage{booktabs} 
\usepackage{makecell}

\usepackage{balance} 

\doi{IVNK5836}

\makeatletter
\gdef\@copyrightpermission{
  \begin{minipage}{0.2\columnwidth}
   \href{https://creativecommons.org/licenses/by/4.0/}{\includegraphics[width=0.90\textwidth]{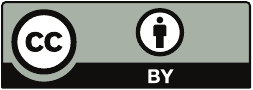}}
  \end{minipage}\hfill
  \begin{minipage}{0.8\columnwidth}
   \href{https://creativecommons.org/licenses/by/4.0/}{This work is licensed under a Creative Commons Attribution International 4.0 License.}
  \end{minipage}
  \vspace{5pt}
}
\makeatother

\setcopyright{ifaamas}
\acmConference[AAMAS '26]{Proc.\@ of the 25th International Conference
on Autonomous Agents and Multiagent Systems (AAMAS 2026)}{May 25 -- 29, 2026}
{Paphos, Cyprus}{C.~Amato, L.~Dennis, V.~Mascardi, J.~Thangarajah (eds.)}
\copyrightyear{2026}
\acmYear{2026}
\acmDOI{}
\acmPrice{}
\acmISBN{}

\title[AAMAS-2026 Formatting Instructions]{Puzzle it Out: Local-to-Global World Model for Offline Multi-Agent Reinforcement Learning}

\author{Sijia Li}
\affiliation{
  \institution{The Hong Kong University of Science and Technology}
  \city{Hong Kong}
  \country{China}}
\email{slifg@connect.ust.hk}

\author{Xinran Li}
\affiliation{
  \institution{The Hong Kong University of Science and Technology}
  \city{Hong Kong}
  \country{China}}
\email{xinran.li@connect.ust.hk}

\author{Shibo Chen}
\authornote{Corresponding authors.} 
\affiliation{
  \institution{South China University of Technology}
  \city{Guangdong}
  \country{China}}
\email{shibochen.ustc@gmail.com}

\author{Jun Zhang}
\authornotemark[1] 
\affiliation{
  \institution{The Hong Kong University of Science and Technology}
  \city{Hong Kong}
  \country{China}}
\email{eejzhang@ust.hk}

\begin{abstract}
Offline multi-agent reinforcement learning (MARL) aims to solve cooperative decision-making problems in multi-agent systems using pre-collected datasets. Existing offline MARL methods primarily constrain training within the dataset distribution, resulting in overly conservative policies that struggle to generalize beyond the support of the data. While model-based approaches offer a promising solution by expanding the original dataset with synthetic data generated from a learned world model, the high dimensionality, non-stationarity, and complexity of multi-agent systems make it challenging to accurately estimate the transitions and reward functions in offline MARL. Given the difficulty of directly modeling joint dynamics, we propose a local-to-global (LOGO) world model, a novel framework that leverages local predictions—which are easier to estimate—to infer global state dynamics, thus improving prediction accuracy while implicitly capturing agent-wise dependencies. Using the trained world model, we generate synthetic data to augment the original dataset, expanding the effective state-action space. To ensure reliable policy learning, we further introduce an uncertainty-aware sampling mechanism that adaptively weights synthetic data by prediction uncertainty, reducing approximation error propagation to policies. In contrast to conventional ensemble-based methods, our approach requires only an additional encoder for uncertainty estimation, significantly reducing computational overhead while maintaining accuracy. Extensive experiments across 8 scenarios against 8 baselines demonstrate that our method surpasses state-of-the-art baselines on standard offline MARL benchmarks, establishing a new model-based baseline for generalizable offline multi-agent learning.
\end{abstract}


\newcommand{\BibTeX}{\rm B\kern-.05em{\sc i\kern-.025em b}\kern-.08em\TeX}

\begin{document}


\pagestyle{fancy}
\fancyhead{}


\maketitle 


\section{Introduction}

Multi-Agent Reinforcement Learning (MARL) tackles scenarios where multiple agents learn and interact concurrently to solve cooperative tasks with a shared goal. Its ability to model complex dynamics and enable joint optimization has led to widespread applications, ranging from robotics to games and distributed systems \cite{gu2023safe, lee2022marl, lanctot2017unified, noguer2024game, lin2020distributed, zhang2025multi}. However, online MARL training faces several critical challenges, including sample inefficiency and non-stationarity, arising from its continuous interaction with the environment. This results in high computational costs, poor sample utilization, and safety concerns in online exploration, thereby hindering its deployment in critical real-world applications such as autonomous driving and clinical decision support \cite{zhang2024multi, jiang2024multi, lin2024multi}. 

In contrast, offline MARL \cite{formanek2024dispelling} trains agents using pre-collected datasets, eliminating the need for active environment interaction. It improves sample efficiency and stability by learning from fixed dataset, avoiding the costs and risks associated with online exploration. However, offline MARL methods face out-of-distribution (OOD) issues, as the fixed dataset cannot account for unseen states or actions. To address this difficulty, current offline MARL approaches employ conservative value estimation \cite{wang2024offline, shao2024counterfactual} and pessimistic policy training \cite{liu2025offline, pan2022plan} by restricting policies to the dataset support. While effective in mitigating OOD errors, these model-free methods often result in overly conservative behaviors, limiting the optimality of learned policies.

\begin{figure}
\centering
\centerline{\includegraphics[width=0.5\textwidth]{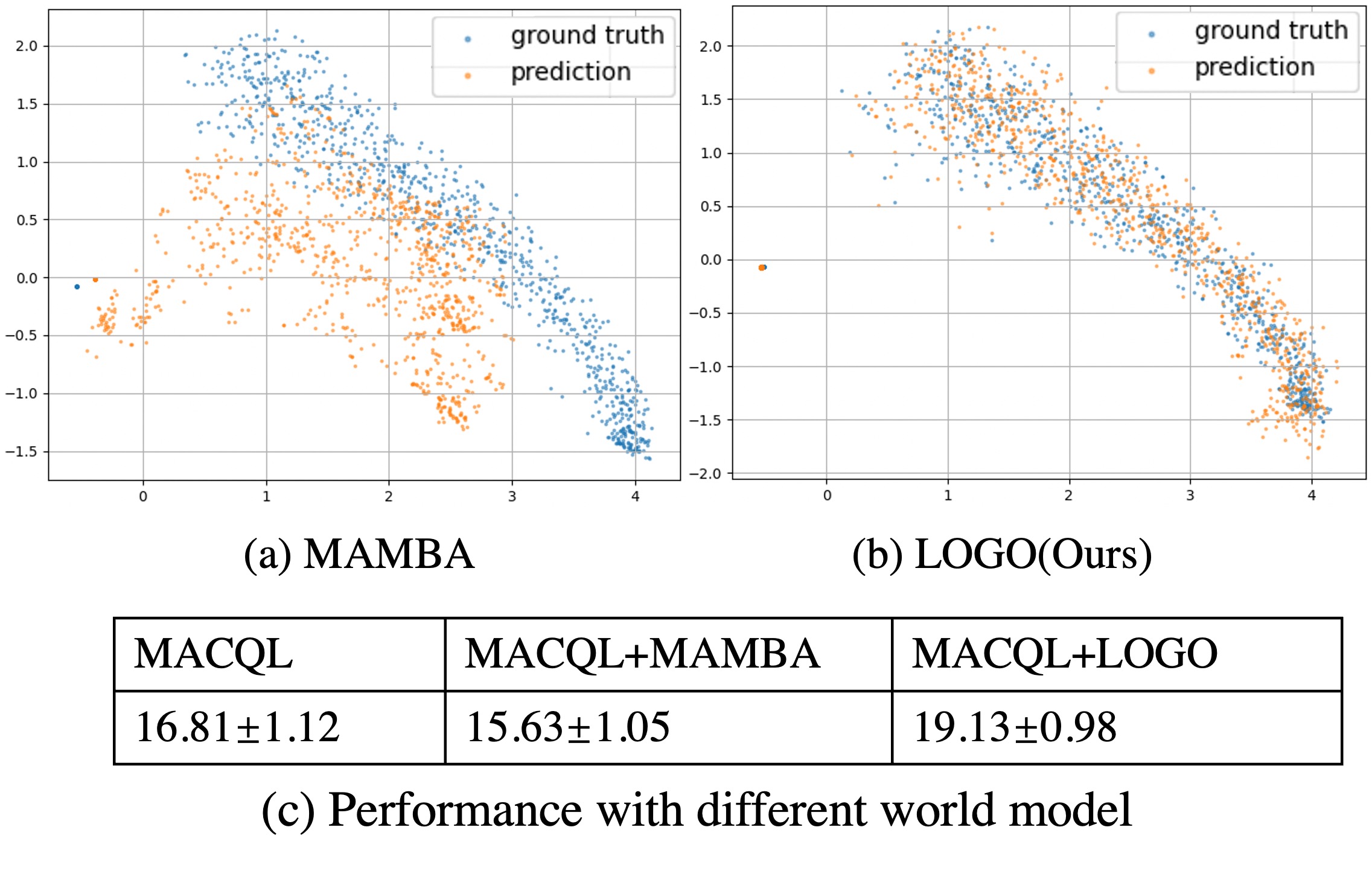}}
\caption{ \textbf{The motivation of LOGO.} The results in (a) and (b) demonstrate the next state prediction accuracy of baseline MAMBA and our proposed LOGO method on the SMAC 5m\_vs\_6m scenario, while (c) presents the performance of MACQL (Multi-Agent Conservative Q-Learning) when integrated with different world models. These results indicate that employing an inaccurate world model can adversely affect overall performance. }
\label{fig:motivation}
\vskip -0.1in
\end{figure}

Model-based methods provide a potentially promising solution to alleviate this over-conservatism by expanding the offline dataset with policy rollouts in world models. However, constructing accurate world models \cite{barde2023model} for offline MARL introduces significant challenges. First, the inherent interaction complexity and high-dimensionality of multi-agent systems brings excessive computational demands and difficulties in learning accurate dynamics with limited data \cite{huang2024modeling}. Second, these difficulties inevitably induce approximation errors in the learned dynamic model, which can propagate during policy rollouts, degrading policy performance. As illustrated in Fig. \ref{fig:motivation}, the baseline world model produces inaccurate transition estimations which negatively impact downstream policy learning. These challenges necessitate both more precise multi-agent dynamic modeling and effective management of model-generated data to enhance generalization while ensuring training stability.

To address these limitations, this paper develops a Local-to-Global (LOGO) world model that improves multi-agent dynamic modeling by leveraging local transitions to assist the global transition prediction. Specifically, LOGO first learns local dynamics models from each agent's observational space, enabling more accurate and efficient learning due to the reduced dimensionality and focusing on individual dynamics. The local predictions are then integrated to infer global state dynamics, piecing them together like a puzzle and capturing inter-agent dependencies and bridging the gap between local and global system behaviors. Furthermore, to account for the inevitable approximation errors in our model, we add an auxiliary state encoder that efficiently estimates prediction uncertainties using the discrepancy between different prediction paths with lower computational cost than conventional ensemble-based methods, enhancing the world model's robustness.

We summarize our contributions as follows:
\begin{itemize}
\item We propose a novel Local-to-Global world model (LOGO) that first captures local observation dynamics then integrating them to deduce the global state, which enables more efficient and accurate modeling of multi-agent transitions and interactions while preserving stability.
\item With LOGO, we augment the original offline dataset with synthetic data with uncertainty-aware weighted sampling to avoid model errors propagating to policy learning. Specifically, we propose a lightweight uncertainty estimation mechanism based on prediction discrepancy across different paths. This approach achieves robust error estimation at significantly lower computational cost than traditional ensemble methods. \item Empirical results on established offline MARL benchmarks demonstrate that our method achieves superior performance compared to existing baselines, highlighting its effectiveness in improving generalization and policy robustness.
\end{itemize}


\section{Related Work}
\label{related_work}

\paragraph{Offline MARL} Offline MARL optimizes policies using pre-collected datasets, removing the need for environment interactions during training and relaxing the requirement for expert demonstrations \cite{yang2021believe}. Recent advances in offline MARL have leveraged diverse methodological approaches to address the challenges of offline learning. From a multi-agent modeling perspective, value-based methods have been enhanced through counterfactual credit assignment \cite{shao2024counterfactual}, local-to-global value decomposition \cite{wang2024offline}, causal inference frameworks \cite{wang2023macca}, and improved temporal credit assignment \cite{eldeeb2024conservative,tian2023learning,yuan2025efficient}. Concurrently, data-centric innovations employ advanced architectures like transformers for local-global policy guidance \cite{tseng2022offline} and diffusion models for trajectory generation \cite{zhu2024madiff,huang2024diffusion,li2023beyond,lidof}, while novel data augmentation techniques address dataset limitations \cite{meng2024new}. On the policy optimization front, methods range from strict in-distribution behavior constrain \cite{liu2025offline,matsunaga2023alberdice} to approaches encouraging controlled diversity within dataset constraints \cite{pan2022plan,huang2024multi,zhou2025cooperative}, with regularization techniques enabling multi-objective optimization \cite{zhan2024exploiting,bui2024comadice}. These advances demonstrate progress in tackling offline MARL challenges through complementary value estimation, data utilization, and policy constraint strategies. However, model-free methods often produce overly conservative policies limited to dataset support, motivating model-based approaches that can generate synthetic state-action pairs via pessimistic rollouts while maintaining stability and encouraging generalization.

\paragraph{Model-based MARL} Model-based MARL employs learned dynamics models to simulate agent interactions for sample-efficient policy optimization. Current approaches primarily address key multi-agent challenges, including scalability \cite{egorov2022scalable}, decentralized local observations \cite{wu2023models}, and latent world models for state-space compression. Specifically, architectures like Dreamer \cite{hafner2023mastering, toledo2024codreamer, lobos2022ma} and Transformer-based models \cite{zhang2022relational} employ latent state encoding to autoregressively predict future states, enabling high-fidelity trajectory simulation and efficient policy training. Online MARL methods \cite{sessa2022efficient, zhang2023model, pasztor2021efficient, huang2024model, xu2022mingling} focus on exploration and sample efficiency, while offline model-based approaches address OOD generalization, requiring careful balance between leveraging learned dynamics for synthetic data and avoiding over-reliance on uncertain extrapolations beyond the dataset support. Several online MARL studies have also adopted a local–global perspective to achieve more accurate prediction, such as MACD \cite{chai2024aligning} and MABL \cite{venugopal2023mabl}. However, these approaches rely on large amounts of data and continuous environment feedback to refine their world models, which makes them unsuitable for straightforward extension to offline MARL. Our work advances this in offline MARL by developing local-to-global world models with enhanced accuracy and incorporating them into policy optimization through an uncertainty-weighting mechanism.

\paragraph{Offline model-based RL} Model-based offline RL employs a learned dynamics model to augment the dataset and improve generalization, which have demonstrated superior performance compared to model-free approaches in offline settings \cite{swazinna2022comparing, yu2020mopo, kidambi2020morel}. The standard pipeline involves jointly training a probabilistic dynamics model and reward function, enabling synthetic trajectory generation through model rollouts. However, due to inevitable approximation errors, careful uncertainty quantification becomes essential - most commonly implemented through ensemble variance \cite{yu2020mopo, kidambi2020morel,rigter2022rambo, lin2022model, wu2024ocean, kidambi2021mobile}, while other methods such as count-based methods \cite{kim2023model} and Bellman estimations \cite{sun2023model} have also demonstrated effectiveness. Alternatively, distributional \cite{shi2024distributionally} or adversarial training paradigms \cite{lin2020model} offer viable approaches to learn the world model. To ensure policy robustness against model errors, existing methods typically incorporate conservatism at either the policy optimization stage \cite{chemingui2024offline, zhu2024model, hishinuma2021weighted, uehara2021pessimistic} or value estimation level \cite{yu2021combo, rigter2022rambo}. In this vein, \citet{barde2023model} introduces a model-based offline MARL algorithm featuring the inter-agent coordination. While advances in single-agent model-based offline RL such as uncertainty-aware learning \cite{diehl2021umbrella} and conservative regularization \cite{liu2023domain} show promise, their ability to extend to offline MARL remains limited by the difficulty of modeling complex interactions among multiple agents and the excessive computational costs of ensembling such models.

\section{Preliminaries}
\label{sec:Preliminaries}

\paragraph{Offline MARL} In this work, we adopt the decentralized partially observable Markov decision process (Dec-POMDP) \cite{oliehoek2016concise} framework for cooperative MARL, formally defined by the tuple $ M = \langle S, A, P, R, \Omega, O, n, \gamma \rangle$, where $n$ agents learn cooperative policies under partial observability. Each agent $i$ receives only local observation \(o^i \in O\) based on the observation function \( Z(s, i) : S \times N \to O \), and selects actions \(a^i \in A\) according to its policy \(\pi^i\). The environment evolves through joint actions \(\boldsymbol{a} = (a^1,...,a^n)\) according to transition dynamics \(P(s' \mid s, \boldsymbol{a}) : S \times A^n \times S' \to [0, 1]\), while providing a shared global reward \( r = R(s, \boldsymbol{a}) \in \mathbb{R} \) to guide policy training. The discount factor \(\gamma \in [0, 1)\) determines the agent's temporal preference by exponentially weighting future rewards in the return calculation \( \mathbb{E}\left[\sum_{t=0}^{\infty} \gamma^t r_t\right]\). Offline MARL algorithms aim to derive an optimal joint policy \( \pi = (\pi^1, \pi^2...,\pi^n)\) from a static dataset consisting of multi-agent transition tuples, where the data is generated by the unknown behavioral policies.

\paragraph{Offline model-based RL} In offline model-based RL, we learn a dynamics model $\hat{M} = (\hat{T}, \hat{r})$ from a fixed dataset $D = \big\{(s, a, s', r)\big\}^N_{i=1}$, where $\hat{T}(s'|s,a)$ is the estimated transition function and $\hat{r}(s,a)$ is the estimated reward function. The model generates synthetic transitions $(s,a,s',r)$ to form a training dataset $\mathcal{D}_m$ for policy optimization, where $s' \sim \hat{T}(s'|s,a)$ and $r \sim \hat{r}(r|s,a)$. To alleviate model errors, a common approach is to employ uncertainty-penalized rewards, which discourage the policy from exploiting regions with high model uncertainty and thereby mitigate the risk of performance degradation induced by model bias or approximation errors. The reward penalty is defined as:
\begin{equation}
    \tilde{r}(s,a) = \hat{r}(s,a) - \lambda u(s,a)
\end{equation}
where $u(s,a)$ estimates state-action uncertainty (typically via model ensembles) and $\lambda$ controls conservatism.

\section{Method}
\label{Method}

\begin{figure*}[ht] 
\vskip -0.1in
\centering
\centerline{\includegraphics[width=0.95\textwidth]{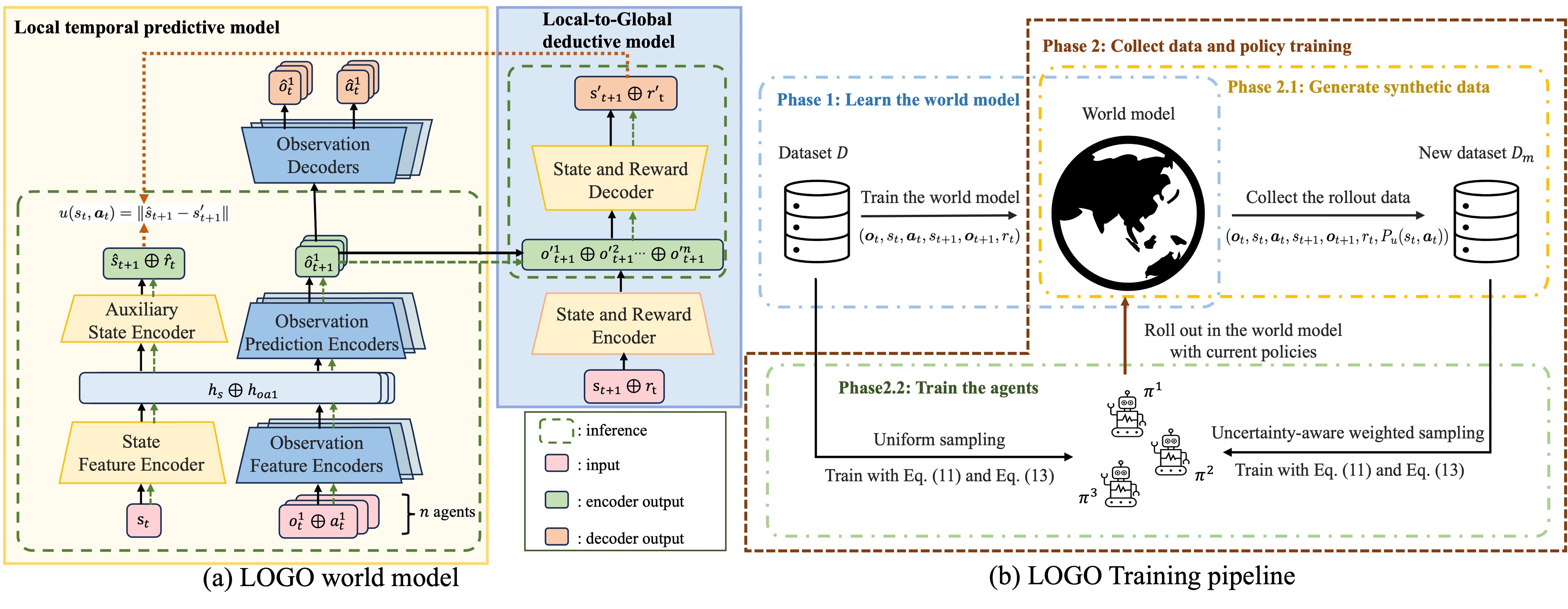}}
\caption{ \textbf{Overall framework of LOGO.} (a) We first train the LOGO world model with both prediction loss and reconstruction loss. Subsequently, during the dynamic data generation phase, the predictive model (yellow part) predicts $\boldsymbol{\hat{o}}_{t+1}$ with ($\boldsymbol{o}_t$, $\boldsymbol{a}_t$, $s_t$). These predictions are then propagated through the deductive model's (blue part) decoder (indicated by the green dashed pathway) to compute the subsequent state $s'_{t+1}$ and reward $r'_t$. Simultaneously, the state outputs from the uncertainty layer $\hat{s}_{t+1}$ along with the deduced state $s'_{t+1}$ are utilized to estimate the prediction uncertainty (indicated by the orange dashed pathway). (b) For policy training, we sample from the generated data using uncertainty-aware weighting to prioritize more reliable transitions.}
\label{fig:framework}
\end{figure*}



As illustrated in Fig. \ref{fig:motivation}, excessive model errors in the learned world model can significantly degrade the performance of offline MARL. Motivated by this observation (Sec. \ref{sec:motivation}), we introduce two key innovations in our proposed LOGO: 1) a more accurate multi-agent world model architecture that reconstructs global states through precise local dynamic predictions (Sec. \ref{sec:world_mode_construct}), and 2) an uncertainty estimation mechanism that quantifies prediction uncertainty in the transition and reward functions.  (Sec. \ref{sec:uncertainty}). Leveraging this uncertainty-aware world model, we generate additional high-quality synthetic data to augment policy training through uncertainty weighted sampling, thereby improving the agents' generalization capability and overall performance.

\subsection{Motivation Example} \label{sec:motivation}

Current model-based MARL methods predominantly rely on direct state prediction to learn world dynamics. However, as illustrated in Fig. \ref{fig:motivation}, model-based MARL methods following this paradigm - including MAMBA - suffer from inherent limitations in state prediction accuracy due to the compounding challenges of multi-agent systems, including partial observability, non-stationary dynamics, and high-dimensional joint action spaces. Such inaccuracies in world modeling further propagate through policy training, ultimately degrading the quality of the learned policy and the performance, as demonstrated in Fig. \ref{fig:motivation}.(c).

To address this issue, we propose a novel world model framework that leverages local prediction as a more tractable intermediate representation. Our key insight is that local observations (e.g., agent-centric states or neighborhood information) can be predicted with significantly higher fidelity than global states, as they circumvent the curse of dimensionality inherent in full-state modeling. By deducing the global state with these more reliable local predictions, our approach achieves more accurate global state estimation while maintaining computational efficiency. This hierarchical local-to-global paradigm not only mitigates error accumulation but also enables adaptive focus on critical interaction patterns within the multi-agent system.


\subsection{Local-to-global World Model Construction}
\label{sec:world_mode_construct}

The key idea of our proposed LOGO world model is to bypass the curse of dimensionality typical in model-based MARL and obtain a more accurate world model by leveraging local dynamic prediction as an intermediate representations to reconstruct the global state, as illustrated in Fig. \ref{fig:framework}. We model each agent independently, where their observations naturally contain information about nearby agents and the environment. This allows us to train a local predictive model that focuses only on each agent's immediate dynamics, which are simpler and more stable compared to modeling the full system. Instead of directly predicting complex global state transitions, we first predict each agent's next local observation and then combine these predictions to reconstruct the complete global state - similar to assembling pieces of a puzzle. This approach improves estimation accuracy through independently modeling agent-level dynamics while leveraging environmental context to implicitly capture inter-agent dependencies. By decoupling local prediction from the unstable and high-dimensional global prediction, the method reduces error propagation and enhance computational efficiency. The local-to-global abstraction aligns with modular MARL architectures, empirically improving training stability and generalization.

As shown in Fig. \ref{fig:framework} (a), our proposed LOGO world model consists of a local world model (predictive model for each agent) illustrated in yellow background, and a deductive model illustrated in blue background. The training path is illustrated by solid lines, whereas the inference path is represented by dashed lines. 

\textbf{During the training phase, }the predictive model first takes $(o^i_t, a^i_t)$ as input for each agent $i$ and use state information $s_t$ as auxiliary information to predict the $\hat{o}^i_{t+1}$. The objective is to enable the encoder part of the local predictive model to accurately predict the next local observation. The deductive model will take the next state $s_{t+1}$ and the reward $r_t$ as input to construct the next joint observations $(\hat{o}^1_{t+1}, \hat{o}^2_{t+1}, ..., \hat{o}^n_{t+1})$. It then uses these observations to deduce a revised next state $s'_{t+1}$ and the reward $r'_t$, thereby enabling the decoder to reconstruct the global state from local observations. For implementation, we initiate both local predictive models and deductive model as auto-encoder structures as they could capture critical dynamic features in a self-supervised manner for more robust representation. This local-to-global approach ensures that the model leverages both local agent interactions and global state information for more accurate and robust prediction. The predictive model’s output is denoted by the hat symbol $\hat{}$, while the output of the deductive model is indicated by the prime notation $'$.

\textbf{During inference,} the local predictive model first encodes the current state $s_t$ via a state feature encoder to get state representation $h_{s}$, while agent-specific observations $o^i_t$ and action $a^i_t$ are processed through local observation feature encoders to obtain representation $h_{oai}$. The observation predictive encoders then generate per-agent next-observation estimates $\hat{o}_{t+1}$. These predictions are subsequently concatenated as $(\hat{o}^1_{t+1}, \hat{o}^2_{t+1}...\hat{o}^n_{t+1})$ and are processed within the decoder module of the deductive model to infer the next global state ${s}'_{t+1}$ and estimated reward $r'$. 

\paragraph{Local predictive model.} During the world model training stage, we minimize the prediction error and the reconstruction error for the predictive model with the loss given as: 
\begin{multline}\label{predictive:combined}
L_{\text{p}} = \underbrace{\mathbb{E}_{(o^i,s_t,a_t,o^i_{t+1})\sim\mathcal{D}}\Big[-\log \widehat{T}(o^i_{t+1}|s_t,o^i_t,a^i_t)\Big]}_{\text{encoder loss}} \\ + \underbrace{\mathbb{E}_{\hat{o}^i_{t+1} \sim \mathcal{E}_p({o}^i_t,a^i_t,s_t)}\Big[\|[o^i_t \oplus a^i_t]-\mathcal{R}_p(\hat{o}^i_{t+1})\|^2\Big]}_{\text{decoder loss}},
\end{multline}
where $\mathcal{E}_p$ and $\mathcal{R}_p$ are the encoder and decoder of the predictive model and $\widehat{T}$ is dynamics model trained with maximum likelihood estimation. The encoder predict $\hat{o}_{t+1}$ and the decoder output $\hat{o}_{t}$ and $\hat{a}_{t}$. Here \(\oplus\) denotes the vector concatenation operation. The decoder incorporates an auxiliary reconstruction loss, which is the second tern in Eq.~\eqref{predictive:combined}, to stabilize world model training and boost representational capacity.

\paragraph{Local-to-Global deductive model.} After obtaining accurate observation predictions $\hat{o}^i_{t+1}$ from the encoder of each local predictive model, we construct the global state by aggregating these predictions into a joint observation $(\hat{o}^1_{t+1}, \hat{o}^2_{t+1}...\hat{o}^n_{t+1})$. This phase, referred to as the deduction stage, ensures a coherent integration of local predictions into a global context.

To further enhance the robustness modeling state-observation relationships and capture the underlying information, we also adopt an autoencoder-based framework in which the encoder takes the next state and reward $(s_{t+1}, r_t)$ as input and learns to reconstruct them as training objectives. Meanwhile, the aggregated observations derived from the predictive model serve as the decoder's input during the training phase. The loss functions of the deductive model are given as:
\begin{multline} \label{deduce:combined}
L_{\text{d}} = \underbrace{\mathbb{E}_{(s_{t+1},r_t,o^i_{t+1})\sim\mathcal{D}}\Big[-\log \widehat{P}(o^i_{t+1}|s_{t+1},r_t)\Big]}_{\text{encoder loss}} \\ + \underbrace{\mathbb{E}_{{o}'^i_{t+1} \sim \mathcal{E}_d(s_{t+1},r_t)}\Big[\|[s_{t+1}\oplus r_t]-\mathcal{R}_d(o'^i_{t+1})\|^2\Big]}_{\text{decoder loss}},
\end{multline}
where $\mathcal{E}_d$ and $\mathcal{R}_d$ are the encoder and decoder of the deductive model and $\widehat{P}$ is a probability function with maximum likelihood estimation. 

To further enhance the model’s deductive capability, we introduce an additional loss term that regularizes the deductive model to improve its inference accuracy on predicted output:
\begin{align} \label{deduce:decoder_2}
L_{\mathcal{R}_d}=\mathbb{E}_{\hat{o}^i_{t+1} \sim {\mathcal{E}_p}({o}^i_t,a^i_t,s_t)}\Big[\|s_{t+1}\oplus r_t]-\mathcal{R}_d(\hat{o}^i_{t+1})\|^2\Big],
\end{align}
where $\mathcal{E}_p$ represents the predictive model’s encoder. The decoder of the deductive model is therefore trained with both Eq.~\eqref{deduce:decoder_2} and Eq.~\eqref{deduce:combined}.

The overall loss function of the world model is defined as the sum of the losses of each component:
\begin{align*} \label{function:world}
L_{world}= L_{\text{p}}+L_{\text{d}}+L_{\mathcal{R}_d}.
\end{align*}

\subsection{Weighted Sampling with Uncertainty Estimation}
\label{sec:uncertainty}
As demonstrated in Fig. \ref{fig:motivation}, inaccurate world model predictions adversely affect policy learning performance through synthetic data generated by the world model that contains significant approximation errors. A common method to address this problem is reward penalty, as mentioned in Sec. \ref{sec:Preliminaries}. However, if the reward function itself is underestimated, applying a penalty could further degrade performance. To mitigate this issue, we propose an uncertainty-based reweighted sampling strategy. This approach prioritizes synthetic trajectories generated by the world model that exhibit high prediction confidence, while down-weighting those associated with high uncertainty.

\paragraph{Uncertainty estimation} To achieve this uncertainty-aware reweighting, we implement an uncertainty estimation module that quantifies prediction confidence by measuring the discrepancy between state-reward $s_{t+1}, r_t$ predictions from different computational paths (the predictive model and deductive model as shown in Fig. \ref{fig:framework}). This structural divergence between models naturally captures prediction variability, yielding a principled measure of epistemic uncertainty. The uncertainty is defined as:
\begin{align}
    u(s_t,\boldsymbol{a}_t)= \|\hat{s}_{t+1}-s'_{t+1}\|,
\end{align}
where $\hat{s}_{t+1}$ is the state output of the auxiliary state encoder in the predictive model and $s'_{t+1}$ is the state output in the deductive model. The auxiliary state encoder that assist the uncertainty estimation above (detailed in Fig. \ref{fig:framework}) is trained with the auxiliary loss given as:
\begin{align} \label{predictive:uncertainty}
L_{\mathcal{E}_{ps}}=\mathbb{E}_{(o^i,s_t,a_t,s_{t+1},r_t)\sim\mathcal{D}}\Big[-\log \widehat{T}(s_{t+1},r_t|s_t,o^i_t,a^i_t)\Big],
\end{align}
where $\mathcal{E}_{ps}$ is the uncertainty layer in the predictive model and $\widehat{T}$ is the dynamics model trained with maximum likelihood estimation.

\paragraph{Weighted sampling}This constrained uncertainty metric is then utilized as a sampling weight during data selection from the world model's generated dataset $D_m$. Specifically, when a predicted experience $(s_{t+1},r_t)$ exhibits higher uncertainty, we assign it a correspondingly lower sampling probability in $D_m$. This weighting scheme prioritizes high-confidence predictions during policy training, mitigating the impact of uncertain or inaccurate model-generated transitions to prevent training degradation.

In our framework, each synthetic experience is stored as 
\[(\boldsymbol{o}_t,s_t,\boldsymbol{a}_t,s_{t+1},\boldsymbol{o}_{t+1},r_t,P_u(s_t,\boldsymbol{a}_t)),\] where $P_u(s_t,\boldsymbol{a}_t) = \text{clip}(C-u(s_t,\boldsymbol{a}_t),[0,C])$ is defined as the clipped version with constant $C$, ensuring valid bounds while prioritizing high-confidence (low-uncertainty) samples during training. While we uniformly sample from the original dataset $D$ to preserve its distribution, we apply uncertainty-weighted sampling exclusively to world model-generated dataset $D_m$. The sampling weight for $D_m$ is inversely proportional to its associated uncertainty is defined as:
\begin{align} \label{uncertainty_weight}
w(s_t,\boldsymbol{a}_t)=\frac{exp(P_u(s_t,\boldsymbol{a}_t))}{\sum_{s_i,a_i \in D_m}exp(P_u(s_i,\boldsymbol{a}_t))} .
\end{align}
This weighting scheme reduces the influence of high-uncertainty generated experience during policy optimization, improving training robustness and mitigating policy degradation due to prediction error.

Under our proposed framework, the overall probability of sampling data from the generated dataset $D_m$ is maintained equal to the overall probability of sampling from the true dataset $D$. Formally, if we denote the optimal Q-value function estimated under uniform sampling as
\begin{align}
    Q^*(s_t, \boldsymbol{a}_t) = \mathbb{E} \left[ r(s_t, \boldsymbol{a}_t) + \gamma \max_{\boldsymbol{a}_{t+1}} Q^*(s_{t+1}, \boldsymbol{a}_{t+1}) \right],
\end{align}
the optimal Q-value under uncertainty-aware weighted sampling estimation can be expressed as:
\begin{align}
    Q^*(s_0, a_0) = \mathbb{E} \left[ w(s_t,\boldsymbol{a}_t)[r(s_t, \boldsymbol{a}_t) + \gamma \max_{\boldsymbol{a}_{t+1}} Q^*(s_{t+1}, \boldsymbol{a}_{t+1})] \right],
\end{align}
where $\mathbb{E}[w(s_t,\boldsymbol{a}_t)]=1$. We establish the estimation error bound of the optimal Q-function under the generated dataset $D_m$ in Theorem \ref{thm:q_bound} (see Appendix \ref{app:proof} for proof). 

\begin{theorem} \label{thm:q_bound}
With the assumption that Q-function and reward function are Lipschitz continuous functions with Lipschitz constants $L_Q$ and $L_r$, and the assumption that the estimated errors of $r$,$s$ and $Q$ are bounded by $\epsilon_s,\epsilon_r,\epsilon_Q$ respectively, we have the estimation error between generated optimal Q-value $\hat{Q}^*(s, a)$ and true optimal Q-value $Q^*(s, a)$ bounded as:
\begin{align*}
\|\hat{Q}^*(s_t, \boldsymbol{a}_t) - Q^*(s_t, \boldsymbol{a}_t)\| \leq (L_r + \gamma L_Q) \, \epsilon_s + \epsilon_r + \gamma \epsilon_Q.
\end{align*}
\end{theorem}

This mechanism enhances sample efficiency and mitigates the potential negative impact of erroneous model predictions on policy optimization. 

\subsection{Overall Training Framework and Objectives}
\label{sec:implementation}

As illustrated in Fig \ref{fig:framework}.(b), we first train the LOGO world model with losses Eq.~\eqref{predictive:combined}, Eq.~\eqref{deduce:combined}, Eq.~\eqref{deduce:decoder_2} and Eq.~\eqref{predictive:uncertainty}. After pre-training, we iteratively generate additional data with the trained world model and use these synthetic data \( D_m \) alongside the original dataset \( D \) to update the policy with the uncertainty-aware priority $P_u(s_t,\boldsymbol{a}_t)$.

For policy training, our method can be integrated with any existing offline MARL algorithm. In our implementation, we adopt MACQL \cite{kumar2020conservative} as the backbone. The Q-value loss function is:
\begin{multline}
    \label{qvalue}
    \min_Q \alpha \cdot \underbrace{\mathbb{E}_{s \sim \mathcal{D} \cup \mathcal{D}_m} \left[ \mathbb{E}_{s \sim \mathcal{D}, a \sim \pi(a)} \left[  Q_\theta(s,a) \right] - \mathbb{E}_{(s,a) \sim \mathcal{D}} \left[  Q_\theta(s,a) \right] \right]}_{\text{Regularization}} \\
    + \frac{1}{2} \underbrace{\mathbb{E}_{(s,a,r,s') \sim \mathcal{D}} \left[ \left(  Q_\theta(s,a) - \mathcal{T}^\pi  Q_\theta(s,a) \right)^2 \right]}_{\text{Bellman error}},
\end{multline} 
where in the multi-agents setting the Bellman operator $\mathcal{T}$ is defined as:
\begin{align}
    \mathcal{T}^\pi  Q_\theta(s, \boldsymbol{a}) = r(s, \boldsymbol{a}) 
    + \gamma \mathbb{E}_{s' \sim P(\cdot|s,\boldsymbol{a})} \left[ \max_{\boldsymbol{a}'} Q_\theta(s', \boldsymbol{a}' ) \right].
\end{align}

For the policy network optimization, we employ a composite objective function that combines policy gradient maximization with behavior cloning (BC) regularization, formulated as:
\begin{multline} \label{policy}
    \mathcal{L}_{\text{policy}} = \underbrace{\mathbb{E}_{(s,\boldsymbol{a}) \sim \mathcal{D} \cup \mathcal{D}_m} \left[ Q_\theta(s, \boldsymbol{\pi_\phi}(s)) \right]}_{\text{Policy Gradient}} 
    \\ - \lambda \cdot \underbrace{\mathbb{E}_{(o_i,a_i) \sim \mathcal{D} \cup \mathcal{D}_m} \left[ \| \pi_{\phi_i}(o^i) - a^i \|_2^2 \right]}_{\text{BC Regularization}},
\end{multline}
where $\lambda$ is a hyperparameter that controls the trade-off between policy optimization and behavior cloning regularization. We include the pssudo code for the overall training pipeline in Appendix \ref{Pseudocode}.

\begin{table*}[ht]
\small
\centering
\caption{Comparison with Baseline Methods. \textbf{Off/On}:Offline/Online. \textbf{MB/MF}:Model-Based/Model-Free. \textbf{Gl/Lo}:Global/Local. \textbf{UE}: Uncertainty estimation.}
\vskip 0.1in
\label{tab:baselines}
\resizebox{0.95\textwidth}{!}{
\begin{tabular}{lccccp{8cm}}
\toprule
\textbf{Method} & \textbf{Off/On} & \textbf{MB/MF} & \textbf{Gl/Lo} & \textbf{UE} & \textbf{Other Notes} \\
\midrule
OMAR \cite{pan2022plan} & Off & MF & Lo & \text{\ding{53}} & A policy-based method under the decentralized framework \\
MACQL \cite{kumar2020conservative} & Off & MF & Gl & \text{\ding{53}} & An adaption of CQL in the multi-agent domain \\
OMIGA \cite{wang2024offline} & Off & MF & Gl & \text{\ding{53}} & A global-to-local value-based method \\
CFCQL \cite{shao2024counterfactual} & Off & MF & Gl & \text{\ding{53}} & A value-based method with counterfactual credit assignment mechanisms \\
\midrule
Morel \cite{kidambi2020morel} & Off & MB & Lo & \checkmark & An ensemble-based approach with pessimistic value regularization \\
SUMO \cite{qiao2025sumo} & Off & MB & Lo & \checkmark & An ensemble-based approach that estimates uncertainty with search-based method \\
MAMBA \cite{egorov2022scalable} & On & MB & Gl & \text{\ding{53}} & A DreamerV2-based method in MARL \\
MAZero \cite{liu2024efficient} & On & MB & Gl & \text{\ding{53}} & A online method with Monte Carlo Tree Search (MCTS) for policy search \\
\midrule
LOGO(Ours) & Off & MB & Lo\&Gl & \checkmark & A local-to-global world model which explicitly estimates the uncertainty with different prediction paths \\
\bottomrule
\end{tabular}
}
\end{table*}

\section{Experiments}
\label{Experiments}

In this section, we conduct experiments across established offline MARL benchmarks, including the Multi-Agent MuJoCo Environment (MaMuJoCo) \cite{peng2021facmac}, and the StarCraft Multi-Agent Environment (SMAC) \cite{lowe2017multi}. Our experimental dataset follows the same setup as prior works \cite{shao2024counterfactual, wang2024offline}, with different data qualities. The baselines are summarized in Table \ref{tab:baselines}. 
\subsection{Main Results}
\label{Main results}
\begin{table*}[ht]
  \caption{Performance comparison of different algorithms in SMAC. The best result in each setting are highlighted in \textcolor{blue}{blue}. MR: Medium-Replay.}
  \label{tab:experiment_smac}
  \small 
  \centering
  \resizebox{0.95\textwidth}{!}{
    \begin{tabular}{llcccccccc}
    \toprule
    \large Map & \large Data & \large MACQL & \large OMAR & \large CFCQL & \large Morel & \large SUMO & \large MAMBA & \large MAZero & \large LOGO(ours) \\
    \hline
    \multirow{4}{*}{\large 2s3z} 
    & {\large MR} & \makecell{{\large 14.18}{\footnotesize$\pm$1.45}} & \makecell{{\large 15.15}{\footnotesize$\pm$1.45}} & \makecell{{\large 16.10}{\footnotesize$\pm$1.38}} & \makecell{{\large 16.44}{\footnotesize$\pm$1.34}} & \makecell{{\large 16.53}{\footnotesize$\pm$1.40}} & \makecell{{\large 16.37}{\footnotesize$\pm$1.37}} & \makecell{{\large 15.76}{\footnotesize$\pm$1.40}} & \textcolor{blue}{\textbf{\makecell{{\large 17.06}{\footnotesize$\pm$1.33}}}} \\
    & {\large Medium} & \makecell{{\large 13.32}{\footnotesize$\pm$1.37}} & \makecell{{\large 12.68}{\footnotesize$\pm$1.41}} & \makecell{{\large 13.29}{\footnotesize$\pm$1.52}} & \makecell{{\large 13.08}{\footnotesize$\pm$1.37}} & \makecell{{\large 13.17}{\footnotesize$\pm$1.38}} & \makecell{{\large 13.27}{\footnotesize$\pm$1.29}} & \makecell{{\large 13.13}{\footnotesize$\pm$1.34}} & \textcolor{blue}{\textbf{\makecell{{\large 13.82}{\footnotesize$\pm$1.40}}}} \\
    & \large Expert & \makecell{{\large 19.77}{\footnotesize$\pm$0.26}} & \makecell{{\large 19.61}{\footnotesize$\pm$0.29}} & \makecell{{\large 19.93}{\footnotesize$\pm$0.04}} & \makecell{{\large 19.84}{\footnotesize$\pm$0.04}} & \makecell{{\large 19.86}{\footnotesize$\pm$0.04}} & \makecell{{\large 19.77}{\footnotesize$\pm$0.42}} & \makecell{{\large 19.69}{\footnotesize$\pm$0.33}} & \textcolor{blue}{\textbf{\makecell{{\large 19.94}{\footnotesize$\pm$0.20}}}} \\
    & \large Mixed & \makecell{{\large 12.94}{\footnotesize$\pm$1.24}} & \makecell{{\large 13.20}{\footnotesize$\pm$1.14}} & \makecell{{\large 13.21}{\footnotesize$\pm$1.16}} & \makecell{{\large 13.44}{\footnotesize$\pm$1.18}} & \makecell{{\large 13.64}{\footnotesize$\pm$1.13}} & \makecell{{\large 13.54}{\footnotesize$\pm$1.50}} & \makecell{{\large 13.26}{\footnotesize$\pm$1.44}} & \textcolor{blue}{\textbf{\makecell{{\large 14.14}{\footnotesize$\pm$1.35}}}} \\
    \midrule
    
    \multirow{4}{*}{\large 3s\_vs\_5z} 
& \large MR& \makecell{{\large 16.27}{\footnotesize$\pm$1.34}} & \makecell{{\large 14.72}{\footnotesize$\pm$0.91}} & \makecell{{\large 19.07}{\footnotesize$\pm$0.73}} & \makecell{{\large 16.64}{\footnotesize$\pm$1.35}} & \makecell{{\large 17.27}{\footnotesize$\pm$1.29}} & \makecell{{\large 16.82}{\footnotesize$\pm$1.11}} & \makecell{{\large 16.89}{\footnotesize$\pm$1.35}} & \textcolor{blue}{\textbf{\makecell{{\large 19.38}{\footnotesize$\pm$0.98}}}} \\
& \large Medium & \makecell{{\large 18.86}{\footnotesize$\pm$1.20}} & \makecell{{\large 20.09}{\footnotesize$\pm$1.03}} & \makecell{{\large 20.88}{\footnotesize$\pm$1.26}} & \makecell{{\large 18.91}{\footnotesize$\pm$1.39}} & \makecell{{\large 19.51}{\footnotesize$\pm$1.34}} & \makecell{{\large 17.54}{\footnotesize$\pm$1.01}} & \makecell{{\large 2.01}{\footnotesize$\pm$0.66}} & \textcolor{blue}{\textbf{\makecell{{\large 21.48}{\footnotesize$\pm$1.45}}}} \\
& \large Expert & \makecell{{\large 21.03}{\footnotesize$\pm$0.18}} & \makecell{{\large 21.40}{\footnotesize$\pm$0.29}} & \makecell{{\large 21.49}{\footnotesize$\pm$0.15}} & \makecell{{\large 21.12}{\footnotesize$\pm$0.27}} & \makecell{{\large 21.32}{\footnotesize$\pm$0.31}} & \makecell{{\large 21.54}{\footnotesize$\pm$0.52}} & \makecell{{\large 20.65}{\footnotesize$\pm$1.00}} & \textcolor{blue}{\textbf{\makecell{{\large 21.63}{\footnotesize$\pm$0.30}}}} \\
& \large Mixed & \makecell{{\large 17.56}{\footnotesize$\pm$1.22}} & \makecell{{\large 20.03}{\footnotesize$\pm$1.16}} & \makecell{{\large 21.13}{\footnotesize$\pm$1.23}} & \makecell{{\large 19.24}{\footnotesize$\pm$0.96}} & \makecell{{\large 19.40}{\footnotesize$\pm$1.06}} & \makecell{{\large 20.67}{\footnotesize$\pm$1.47}} & \makecell{{\large 19.88}{\footnotesize$\pm$1.12}} & \textcolor{blue}{\textbf{\makecell{{\large 21.43}{\footnotesize$\pm$1.20}}}} \\
\midrule
    
    \multirow{4}{*}{\large 5m\_vs\_6m} 
& \large MR& \makecell{{\large 10.39}{\footnotesize$\pm$1.27}} & \makecell{{\large 10.43}{\footnotesize$\pm$1.37}} & \makecell{{\large 11.35}{\footnotesize$\pm$1.52}} & \makecell{{\large 11.17}{\footnotesize$\pm$1.59}} & \makecell{{\large 11.01}{\footnotesize$\pm$1.61}} & \textcolor{blue}{\textbf{\makecell{{\large 11.82}{\footnotesize$\pm$1.58}}}} & \makecell{{\large 11.12}{\footnotesize$\pm$1.55}} & \makecell{{\large 11.67}{\footnotesize$\pm$1.49}} \\
& \large Medium & \makecell{{\large 9.44}{\footnotesize$\pm$0.96}} & \makecell{{\large 11.84}{\footnotesize$\pm$1.60}} & \makecell{{\large 12.29}{\footnotesize$\pm$1.85}} & \makecell{{\large 11.63}{\footnotesize$\pm$1.32}} & \makecell{{\large 11.82}{\footnotesize$\pm$1.56}} & \makecell{{\large 12.33}{\footnotesize$\pm$1.56}} & \makecell{{\large 10.94}{\footnotesize$\pm$1.53}} & \textcolor{blue}{\textbf{\makecell{{\large 12.63}{\footnotesize$\pm$1.73}}}} \\
& \large Expert & \makecell{{\large 16.35}{\footnotesize$\pm$1.61}} & \makecell{{\large 17.56}{\footnotesize$\pm$1.28}} & \makecell{{\large 17.56}{\footnotesize$\pm$1.51}} & \makecell{{\large 8.77}{\footnotesize$\pm$0.96}} & \makecell{{\large 9.82}{\footnotesize$\pm$0.72}} & \makecell{{\large 16.35}{\footnotesize$\pm$1.61}} & \makecell{{\large 17.01}{\footnotesize$\pm$1.15}} & \textcolor{blue}{\textbf{\makecell{{\large 18.51}{\footnotesize$\pm$1.33}}}} \\
& \large Mixed & \makecell{{\large 7.10}{\footnotesize$\pm$0.58}} & \makecell{{\large 17.56}{\footnotesize$\pm$1.45}} & \makecell{{\large 17.52}{\footnotesize$\pm$1.57}} & \makecell{{\large 13.68}{\footnotesize$\pm$1.88}} & \makecell{{\large 14.23}{\footnotesize$\pm$1.71}} & \makecell{{\large 15.46}{\footnotesize$\pm$1.54}} & \makecell{{\large 15.66}{\footnotesize$\pm$1.51}} & \textcolor{blue}{\textbf{\makecell{{\large 18.21}{\footnotesize$\pm$1.43}}}} \\
\midrule

\multirow{4}{*}{\large 6h\_vs\_8z} 
& \large MR& \makecell{{\large 15.36}{\footnotesize$\pm$1.08}} & \makecell{{\large 12.57}{\footnotesize$\pm$0.74}} & \makecell{{\large 16.13}{\footnotesize$\pm$1.17}} & \makecell{{\large 15.04}{\footnotesize$\pm$1.23}} & \makecell{{\large 15.54}{\footnotesize$\pm$1.16}} & \makecell{{\large 15.65}{\footnotesize$\pm$0.95}} & \makecell{{\large 16.35}{\footnotesize$\pm$0.91}} & \textcolor{blue}{\textbf{\makecell{{\large 17.14}{\footnotesize$\pm$1.51}}}} \\
& \large Medium & \makecell{{\large 11.88}{\footnotesize$\pm$1.04}} & \makecell{{\large 17.24}{\footnotesize$\pm$1.19}} & \makecell{{\large 17.88}{\footnotesize$\pm$1.04}} & \makecell{{\large 17.02}{\footnotesize$\pm$1.12}} & \makecell{{\large 17.34}{\footnotesize$\pm$1.21}} & \makecell{{\large 16.89}{\footnotesize$\pm$1.08}} & \makecell{{\large 17.72}{\footnotesize$\pm$1.30}} & \textcolor{blue}{\textbf{\makecell{{\large 18.70}{\footnotesize$\pm$0.99}}}} \\
& \large Expert & \makecell{{\large 16.81}{\footnotesize$\pm$1.12}} & \makecell{{\large 17.51}{\footnotesize$\pm$0.72}} & \makecell{{\large 18.76}{\footnotesize$\pm$0.85}} & \makecell{{\large 16.78}{\footnotesize$\pm$1.22}} & \makecell{{\large 17.71}{\footnotesize$\pm$1.25}} & \makecell{{\large 15.63}{\footnotesize$\pm$1.05}} & \makecell{{\large 15.90}{\footnotesize$\pm$1.66}} & \textcolor{blue}{\textbf{\makecell{{\large 19.13}{\footnotesize$\pm$0.98}}}} \\
& \large Mixed & \makecell{{\large 12.15}{\footnotesize$\pm$0.84}} & \makecell{{\large 17.36}{\footnotesize$\pm$0.83}} & \makecell{{\large 17.72}{\footnotesize$\pm$0.97}} & \makecell{{\large 17.58}{\footnotesize$\pm$1.13}} & \makecell{{\large 17.81}{\footnotesize$\pm$1.23}} & \makecell{{\large 17.13}{\footnotesize$\pm$1.18}} & \makecell{{\large 17.61}{\footnotesize$\pm$1.05}} & \textcolor{blue}{\textbf{\makecell{{\large 18.68}{\footnotesize$\pm$1.00}}}} \\
\midrule

\multirow{4}{*}{\large 3s5z\_vs\_3s6z} 
& \large MR& \makecell{{\large 8.37}{\footnotesize$\pm$0.54}} & \makecell{{\large 9.64}{\footnotesize$\pm$0.52}} & \makecell{{\large 10.08}{\footnotesize$\pm$0.61}} & \makecell{{\large 10.01}{\footnotesize$\pm$0.60}} & \makecell{{\large 10.03}{\footnotesize$\pm$0.56}} & \makecell{{\large 10.13}{\footnotesize$\pm$0.42}} & \makecell{{\large 9.83}{\footnotesize$\pm$0.67}} & \textcolor{blue}{\textbf{\makecell{{\large 10.37}{\footnotesize$\pm$0.70}}}} \\
& \large Medium & \makecell{{\large 10.23}{\footnotesize$\pm$0.70}} & \makecell{{\large 11.54}{\footnotesize$\pm$0.67}} & \makecell{{\large 11.80}{\footnotesize$\pm$0.61}} & \makecell{{\large 10.49}{\footnotesize$\pm$0.53}} & \makecell{{\large 10.97}{\footnotesize$\pm$0.65}} & \makecell{{\large 11.41}{\footnotesize$\pm$0.73}} & \makecell{{\large 11.35}{\footnotesize$\pm$0.74}} & \textcolor{blue}{\textbf{\makecell{{\large 12.17}{\footnotesize$\pm$0.66}}}} \\
& \large Expert & \makecell{{\large 10.71}{\footnotesize$\pm$0.76}} & \makecell{{\large 11.81}{\footnotesize$\pm$0.74}} & \makecell{{\large 12.05}{\footnotesize$\pm$0.74}} & \makecell{{\large 11.92}{\footnotesize$\pm$0.62}} & \makecell{{\large 11.91}{\footnotesize$\pm$0.74}} & \makecell{{\large 10.81}{\footnotesize$\pm$0.40}} & \makecell{{\large 11.48}{\footnotesize$\pm$0.77}} & \textcolor{blue}{\textbf{\makecell{{\large 12.31}{\footnotesize$\pm$0.76}}}} \\
& \large Mixed & \makecell{{\large 9.87}{\footnotesize$\pm$0.67}} & \makecell{{\large 11.22}{\footnotesize$\pm$0.57}} & \textcolor{blue}{\textbf{\makecell{{\large 11.53}{\footnotesize$\pm$0.56}}}} & \makecell{{\large 10.49}{\footnotesize$\pm$0.63}} & \makecell{{\large 10.65}{\footnotesize$\pm$0.61}} & \makecell{{\large 9.95}{\footnotesize$\pm$0.62}} & \makecell{{\large 10.18}{\footnotesize$\pm$0.51}} & \makecell{{\large 11.12}{\footnotesize$\pm$0.72}} \\
    
    \bottomrule
    \end{tabular}
    }
\end{table*}

\begin{table*}[ht]
  \caption{Performance comparison of different algorithms in MaMuJoCo. The best result in each row are highlighted in \textcolor{blue}{blue}. Ho: Hopper-v2. An: Ant-v2. Ha: HalfCheetah-v2. MR: Medium-Replay. ME: Medium. EX: Expert. MI: Mixed. }
  \label{tab:continous}
  \centering
  \large
  \resizebox{\textwidth}{!}{
    \begin{tabular}{llcccccccc}
    \toprule
    En & Data & MACQL & OMIGA & CFCQL & Morel & SUMO & MAMBA & MAZero & LOGO(ours) \\
    \midrule
    \multirow{4}{*}{Ho} 
    & MR & \makecell{{\Large 330.23}{\footnotesize$\pm$6.23}} & \makecell{{\Large 774.18}{\footnotesize$\pm$197.38}} & \makecell{{\Large 854.23}{\footnotesize$\pm$135.43}} & \makecell{{\Large 1412.36}{\footnotesize$\pm$182.96}} & \makecell{{\Large 1218.48}{\footnotesize$\pm$124.63}} & \makecell{{\Large 350.57}{\footnotesize$\pm$9.33}} & \makecell{{\Large 906.63}{\footnotesize$\pm$102.94}} & \textcolor{blue}{\textbf{\makecell{{\Large 1659.10}{\footnotesize$\pm$130.30}}}} \\
    & ME & \makecell{{\Large 489.23}{\footnotesize$\pm$102.34}} & \makecell{{\Large 1148.62}{\footnotesize$\pm$169.02}} & \makecell{{\Large 1231.34}{\footnotesize$\pm$176.34}} & \makecell{{\Large 976.89}{\footnotesize$\pm$92.26}} & \makecell{{\Large 932.46}{\footnotesize$\pm$106.25}} & \makecell{{\Large 1087.28}{\footnotesize$\pm$140.63}} & \makecell{{\Large 1172.12}{\footnotesize$\pm$146.9}} & \textcolor{blue}{\textbf{\makecell{{\Large 1302.7}{\footnotesize$\pm$189.24}}}} \\
    & EX & \makecell{{\Large 403.26}{\footnotesize$\pm$126.34}} & \makecell{{\Large 872.39}{\footnotesize$\pm$204.27}} & \makecell{{\Large 899.34}{\footnotesize$\pm$186.24}} & \makecell{{\Large 683.77}{\footnotesize$\pm$211.72}} & \makecell{{\Large 823.53}{\footnotesize$\pm$367.35}} & \makecell{{\Large 903.12}{\footnotesize$\pm$375.8}} & \makecell{{\Large 684.72}{\footnotesize$\pm$307.22}} & \textcolor{blue}{\textbf{\makecell{{\Large 1086.44}{\footnotesize$\pm$221.65}}}} \\
    & MI & \makecell{{\Large 376.23}{\footnotesize$\pm$63.23}} & \makecell{{\Large 718.01}{\footnotesize$\pm$285.79}} & \makecell{{\Large 845.23}{\footnotesize$\pm$253.23}} & \makecell{{\Large 1038.11}{\footnotesize$\pm$182.96}} & \makecell{{\Large 1089.32}{\footnotesize$\pm$162.76}} & \makecell{{\Large 303.33}{\footnotesize$\pm$104.67}} & \makecell{{\Large 876.69}{\footnotesize$\pm$198.86}} & \textcolor{blue}{\textbf{\makecell{{\Large 1245.65}{\footnotesize$\pm$244.26}}}} \\
    \midrule

    \multirow{4}{*}{An} 
    & MR & \makecell{{\Large 804.43}{\footnotesize$\pm$83.22}} & \makecell{{\Large 1029.13}{\footnotesize$\pm$21.27}} & \makecell{{\Large 1038.23}{\footnotesize$\pm$43.23}} & \makecell{{\Large 1052.67}{\footnotesize$\pm$22.34}} & \makecell{{\Large 1077.67}{\footnotesize$\pm$25.82}} & \makecell{{\Large 1019.55}{\footnotesize$\pm$26.25}} & \makecell{{\Large 1012.11}{\footnotesize$\pm$16.03}} & \textcolor{blue}{\textbf{\makecell{{\Large 1253.01}{\footnotesize$\pm$18.69}}}} \\
    & ME & \makecell{{\Large 1337.74}{\footnotesize$\pm$56.66}} & \makecell{{\Large 1417.37}{\footnotesize$\pm$4.11}} & \makecell{{\Large 1457.23}{\footnotesize$\pm$7.23}} & \makecell{{\Large 1398.05}{\footnotesize$\pm$5.12}} & \makecell{{\Large 1408.34}{\footnotesize$\pm$5.02}} & \makecell{{\Large 1380.37}{\footnotesize$\pm$4.75}} & \makecell{{\Large 1382.64}{\footnotesize$\pm$4.65}} & \textcolor{blue}{\textbf{\makecell{{\Large 1441.08}{\footnotesize$\pm$6.18}}}} \\
    & EX & \makecell{{\Large 1634.76}{\footnotesize$\pm$125.38}} & \makecell{{\Large 2053.23}{\footnotesize$\pm$3.34}} & \makecell{{\Large 2042.75}{\footnotesize$\pm$4.26}} & \makecell{{\Large 2047.27}{\footnotesize$\pm$3.39}} & \makecell{{\Large 2044.23}{\footnotesize$\pm$2.43}} & \makecell{{\Large 2042.84}{\footnotesize$\pm$2.38}} & \makecell{{\Large 2035.96}{\footnotesize$\pm$4.48}} & \textcolor{blue}{\textbf{\makecell{{\Large 2063.73}{\footnotesize$\pm$2.89}}}} \\
    & MI & \makecell{{\Large 1474.34}{\footnotesize$\pm$86.43}} & \makecell{{\Large 1717.23}{\footnotesize$\pm$37.23}} & \makecell{{\Large 1687.34}{\footnotesize$\pm$58.94}} & \makecell{{\Large 1701.48}{\footnotesize$\pm$29.11}} & \makecell{{\Large 1745.23}{\footnotesize$\pm$32.16}} & \makecell{{\Large 1522.10}{\footnotesize$\pm$48.91}} & \makecell{{\Large 1711.13}{\footnotesize$\pm$41.30}} & \textcolor{blue}{\textbf{\makecell{{\Large 1853.01}{\footnotesize$\pm$32.63}}}} \\
    \midrule
    
    \multirow{3}{*}{Ha} 
    & MR & \makecell{{\Large 2042.69}{\footnotesize$\pm$196.34}} & \makecell{{\Large 2406.70}{\footnotesize$\pm$91.13}} & \makecell{{\Large 2334.87}{\footnotesize$\pm$98.71}} & \makecell{{\Large 2393.57}{\footnotesize$\pm$69.75}} & \makecell{{\Large 2437.57}{\footnotesize$\pm$49.85}} & \makecell{{\Large 2434.29}{\footnotesize$\pm$67.19}} & \textcolor{blue}{\textbf{\makecell{{\Large 2593.38}{\footnotesize$\pm$95.02}}}} & \makecell{{\Large 2566.74}{\footnotesize$\pm$67.63}} \\
    & ME & \makecell{{\Large 1594.34}{\footnotesize$\pm$263.84}} & \makecell{{\Large 2646.26}{\footnotesize$\pm$39.51}} & \makecell{{\Large 2780.06}{\footnotesize$\pm$43.22}} & \makecell{{\Large 2669.34}{\footnotesize$\pm$35.24}} & \makecell{{\Large 2713.48}{\footnotesize$\pm$46.38}} & \makecell{{\Large 2436.24}{\footnotesize$\pm$50.57}} & \makecell{{\Large 2652.67}{\footnotesize$\pm$49.2}} & \textcolor{blue}{\textbf{\makecell{{\Large 2805.25}{\footnotesize$\pm$56.36}}}} \\
    & EX & \makecell{{\Large 2347.13}{\footnotesize$\pm$298.33}} & \makecell{{\Large 3138.54}{\footnotesize$\pm$227.20}} & \makecell{{\Large 3345.80}{\footnotesize$\pm$164.23}} & \makecell{{\Large 3259.01}{\footnotesize$\pm$204.09}} & \makecell{{\Large 3305.89}{\footnotesize$\pm$183.26}} & \makecell{{\Large 2883.30}{\footnotesize$\pm$201.34}} & \makecell{{\Large 3079.40}{\footnotesize$\pm$233.58}} & \textcolor{blue}{\textbf{\makecell{{\Large 3534.30}{\footnotesize$\pm$191.75}}}} \\
    & MI & \makecell{{\Large 2174.25}{\footnotesize$\pm$216.52}} & \makecell{{\Large 2825.58}{\footnotesize$\pm$312.33}} & \makecell{{\Large 2645.23}{\footnotesize$\pm$232.53}} & \makecell{{\Large 2853.10}{\footnotesize$\pm$241.48}} & \makecell{{\Large 2902.35}{\footnotesize$\pm$233.15}} & \makecell{{\Large 2801.10}{\footnotesize$\pm$224.26}} & \makecell{{\Large 2947.46}{\footnotesize$\pm$208.75}} & \textcolor{blue}{\textbf{\makecell{{\Large 3060.46}{\footnotesize$\pm$299.17}}}} \\  
    \bottomrule
    \end{tabular}
    }
\end{table*}

In the experiments, we set the rollout horizon to 15 steps. As demonstrated in Table \ref{tab:rollout_length}, LOGO achieves superior long-term prediction ability compared with existing model-based approaches in the offline setting. All results are presented as mean values with 95\% confidence intervals (mean ± margin of error). The results are shown in Table \ref{tab:experiment_smac} and Table \ref{tab:continous}, which demonstrate that LOGO outperforms both model-based and model-free approaches, particularly with medium-quality datasets. This empirically demonstrates that LOGO achieves superior generalization with superior performance than prior model-free offline methods, while delivering more accurate predictions and more stable policy training compared to model-based offline approaches.



\subsection{Generalization Capability with Model Predictive Control (MPC)} \label{exp:mpc}

\begin{table}[ht]
\centering
\centering
\large
\caption{The performance of MPC.}
\resizebox{0.47\textwidth}{!}{
\begin{tabular}{lcccc}
    \toprule
    \textbf{Method} & \textbf{Med-Rep} & \textbf{Medium} & \textbf{Expert} & \textbf{Mixed} \\
    \midrule
    MACQL & \makecell{{\large 15.36}{\footnotesize$\pm$1.08}} & \makecell{{\large 11.18}{\footnotesize$\pm$1.04}} & \makecell{{\large 12.81}{\footnotesize$\pm$1.12}} & \makecell{{\large 12.15}{\footnotesize$\pm$0.84}} \\
    MACQL+MPC & \makecell{{\large 16.36}{\footnotesize$\pm$0.74}} & \makecell{{\large 17.36}{\footnotesize$\pm$0.63}} & \makecell{{\large 17.98}{\footnotesize$\pm$0.55}} & \makecell{{\large 17.27}{\footnotesize$\pm$0.84}} \\
    \bottomrule
    \end{tabular}
}

\label{tab:MPC}
\end{table}

To validate the generalization capability of our world model, we further implement the LOGO-enabled  Model Predictive Control (MPC) \cite{hansen2022temporal} and  evaluate its effectiveness. We train the MPC-enhanced MARL algorithm using the loss function: \[\mathcal{L}_{\text{MPC}} = \mathcal{L}_{\text{policy}} + \mathbb{E}_{(o_i,a_i) \sim \mathcal{D}} \left[ \| \pi_{\phi_i}(o^i) - a^i_{max} \|_2^2 \right]\], where $a^i_{max}$ represents the optimal action selected through model rollout from three candidate actions proposed by the policy network. Results from the 6h\_vs\_8z map in SMAC are presented in Table \ref{tab:MPC}, showing that LOGO-enabled MPC enjoys significant performance improvements and enhanced stability. This experiments show the adaptivity of the proposed the multi-agent world model LOGO across MBRL algorithms and MPC applications.

\subsection{Efficiency of LOGO} \label{exp:eff}
To evaluate computational efficiency, we conducted extensive benchmarking against ensemble-based world model. Wall-clock time was measured over 500 independent trajectories, each consisting of 10 rollout steps under identical environmental conditions. The results show that our method achieves substantially higher efficiency, with up to a 3× reduction in inference runtime while maintaining comparable accuracy (Table \ref{tab:eff}). These runtime comparisons will be included in the revised manuscript to better highlight the practical advantages of our approach.

\begin{table}[ht]
\centering
\small
\caption{The Comparison of Efficiency.}
\label{tab:eff}
\resizebox{0.35\textwidth}{!}{
\begin{tabular}{lcc}
\toprule
Map & Ensemble & LOGO (ours) \\
\midrule
2s3z       & 0.227 & 0.057 \\
5m\_vs\_6m & 0.225 & 0.042 \\
6h\_vs\_8z & 0.247 & 0.063 \\
\bottomrule
\end{tabular}
}

\end{table}

\begin{table}[ht]
\caption{Ablation study of uncertainty.WS:Weight sampling. RP:Reward penalty.}
  \label{tab:uncertainty}
  \centering
  \large
  \resizebox{0.5\textwidth}{!}{
    \begin{tabular}{lcccc}
    \toprule
    \textbf{Method} & \textbf{Hopper-v2} & \textbf{Ant-v2} & \textbf{HalfCheetah-v2} \\
    \midrule
    WS & \makecell{{\large 1302.34}{\footnotesize$\pm$189.24}} & \makecell{{\large 1436.08}{\footnotesize$\pm$6.18}} & \makecell{{\large 2805.25}{\footnotesize$\pm$56.36}} \\
    RP & \makecell{{\large 1133.46}{\footnotesize$\pm$165.64}} & \makecell{{\large 1245.23}{\footnotesize$\pm$7.45}} & \makecell{{\large 2539.43}{\footnotesize$\pm$74.23}} \\
    \bottomrule
    \end{tabular}
    }
\end{table}

\begin{table}[ht]
\caption{Performance of different rollout horizon and one-step reward prediction error. The best result in each setting are highlighted in \textcolor{blue}{blue}.}
  \label{tab:rollout_length}
  \centering
  \large
  \resizebox{0.5\textwidth}{!}{
    \begin{tabular}{lccccc}
    \toprule
    \textbf{} & \textbf{Morel} & \textbf{SUMO} & \textbf{MAMBA} & \textbf{MAZero} & \textbf{LOGO(ours)} \\
    \midrule
    horizon = 5  & \makecell{{\large 17.56}{\footnotesize$\pm$0.47}} & \makecell{{\large 17.31}{\footnotesize$\pm$0.42}} & \makecell{{\large 17.46}{\footnotesize$\pm$0.54}} & \makecell{{\large 17.14}{\footnotesize$\pm$0.72}} & \textcolor{blue}{\makecell{{\large 18.14}{\footnotesize$\pm$0.46}}} \\
    horizon = 10 & \makecell{{\large 18.02}{\footnotesize$\pm$0.57}} & \makecell{{\large 18.05}{\footnotesize$\pm$0.40}} & \makecell{{\large 18.06}{\footnotesize$\pm$0.57}} & \makecell{{\large 17.14}{\footnotesize$\pm$0.72}} & \textcolor{blue}{\makecell{{\large 18.46}{\footnotesize$\pm$0.56}}} \\
    horizon = 15 & \makecell{{\large 16.78}{\footnotesize$\pm$0.61}} & \makecell{{\large 17.34}{\footnotesize$\pm$0.58}} & \makecell{{\large 16.89}{\footnotesize$\pm$0.54}} & \makecell{{\large 17.72}{\footnotesize$\pm$0.65}} & \textcolor{blue}{\makecell{{\large 18.70}{\footnotesize$\pm$0.49}}} \\
    horizon = 20 & \makecell{{\large 17.23}{\footnotesize$\pm$0.72}} & \makecell{{\large 17.58}{\footnotesize$\pm$0.59}} & \makecell{{\large 16.03}{\footnotesize$\pm$0.48}} & \makecell{{\large 16.03}{\footnotesize$\pm$0.80}} & \textcolor{blue}{\makecell{{\large 18.57}{\footnotesize$\pm$0.58}}} \\
    prediction error &0.1316 & 0.0928 & 0.0673 & 0.0257 & \textcolor{blue}{0.0051} \\
    \bottomrule
    \end{tabular}
  }
  \vskip -0.1in
\end{table}

\subsection{Ablation Study and Visualization Analysis}

\begin{figure*}[ht] 
\centering
\centerline{\includegraphics[width=0.98\textwidth]{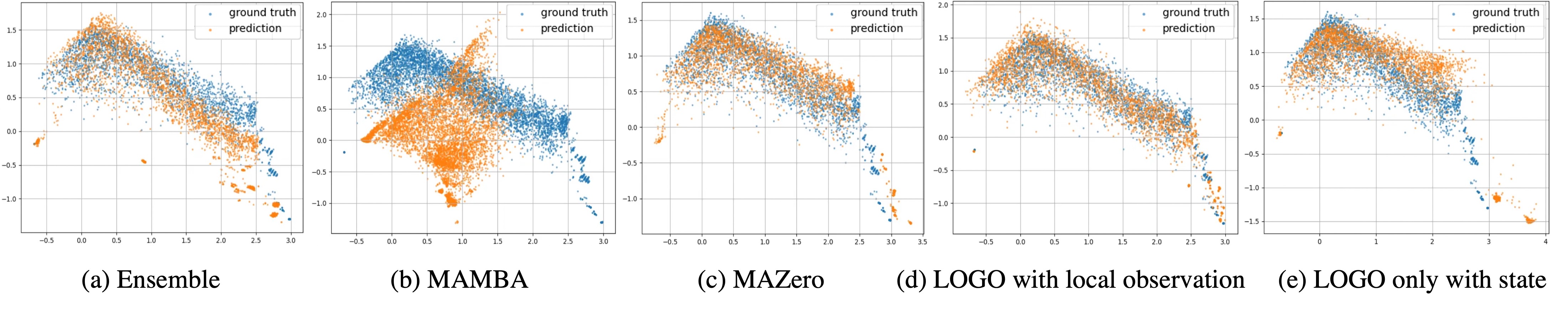}}
\caption{ \textbf{The prediction comparison of model-based methods.} We employ Principal Component Analysis (PCA) \cite{groth2013principal} to visualize and compare the ground truth next state with the predicted next state generated by world models across various model-based approaches in SMAC 6h\_vs\_8z map.  }
\label{fig:vis}
\end{figure*}

\paragraph{Local-to-global Prediction Visualization} \label{exp:able_state} In this experiment, we demonstrate that compared to directly predicting the global state, our method achieves higher accuracy, which models the local dynamics and infers the subsequent state from the predicted observations. We employ the same architecture as the local predictive model to perform direct state prediction. As demonstrated in Fig. \ref{fig:vis} (d) and (e), the proposed approach, which predicts local observations and subsequently infers the system state from these local predictions, yields significantly improved accuracy compared to conventional methods.

\paragraph{Visualization of Prediction Accuracy Comparison} \label{exp:Visualization} We visualize the discrepancy between predicted and ground-truth values with data out of the dataset. The results demonstrate that, compared to other model-based approaches, our framework yields more accurate predictions. The visualization results presented in Fig. \ref{fig:vis} and the one-step reward prediction error reported in the last row of Table \ref{tab:rollout_length} demonstrate that our method achieves more accurate predictions compared to the baseline model-based approaches.

\paragraph{Uncertainty estimation} \label{exp:uncertainty} In offline model-based RL, conventional uncertainty estimation often penalizes rewards to conservatively estimate values and mitigate model errors. Through comparative experiments (Table \ref{tab:uncertainty}) in the medium quality dataset, we demonstrate that our uncertainty-aware sampling outperforms this reward-penalty approach. We identify reward prediction inaccuracies in complex multi-agent settings as a key factor, where reward estimation errors propagate to value approximations - especially with reward shaping. Our uncertainty-weighted sampling addresses this by adaptively prioritizing high-confident samples to reduce estimation bias.

\paragraph{Rollout horizon selection} To assess the effectiveness of different world models using varying trajectory horizons, we conduct comparative experiments using rollout horizons of 5, 10, 15, and 20 steps. As demonstrated in Table \ref{tab:rollout_length}, our proposed LOGO consistently outperforms baseline methods across multiple rollout horizons. Specifically, some baseline methods suffer from performance degradation when using a long rollout horizon (e.g., 20 steps), while ours achieves optimal performance even at 15 steps (compared to their 10-step limit), due to its uncertainty rewighted sampling against error accumulation over extended sequences.


\section{Conclusion}
\label{sec:conclusion}

In this paper, we integrate model-based approaches into offline MARL by introducing a novel Local-to-Global World Model (LOGO) to effectively capture the underlying multi-agent transition dynamics and reward functions. We propose a local-to-global framework capable of capturing precise dynamics from observational space—which is more tractable to learn—and subsequently inferring the global state in a puzzle-like manner to reconstruct the global dynamics with greater accuracy while implicitly modeling inter-agent relationships. LOGO expands the dataset by generating synthetic data based on the learned world model, effectively extending data coverage and improving policy generalization. To further enhance policy reliability, we propose an uncertainty-based weighted sampling mechanism, which adaptively weights synthetic data by prediction uncertainty. This mechanism mitigates the impact of approximation errors on policy learning while maintaining computational efficiency by avoiding the overhead of traditional ensemble-based uncertainty estimation methods. 



\bibliographystyle{ACM-Reference-Format} 
\bibliography{sample}


\end{document}